\documentclass{article}

\usepackage[utf8]{inputenc}
\usepackage[T1]{fontenc}
\usepackage{hyperref}
\usepackage{url}
\usepackage{booktabs}
\usepackage{amsfonts}
\usepackage{nicefrac}
\usepackage{microtype}

\usepackage{graphicx}
\usepackage{subfig}
\usepackage{float}
\usepackage{wrapfig}

\usepackage{algorithm}
\usepackage{algorithmic}
\usepackage{amssymb}
\usepackage{amsmath}
\usepackage{mathtools}
\usepackage{amsthm}
\usepackage{bbm}
\usepackage{makecell}

\DeclarePairedDelimiter{\ceil}{\lceil}{\rceil}

\newtheorem{condition}{Condition}
\newtheorem{assumption}{Assumption}
\newtheorem{theorem}{Theorem}

\newtheorem{corollary}[theorem]{Corollary}

\newcommand{\E}{\mathop{\mathbb{E}}}
\newcommand{\one}{\mathbbm{1}}
\newcommand{\R}{\mathbbm{R}}
\newcommand{\X}{\mathcal{X}}
\newcommand{\Y}{\mathcal{Y}}
\newcommand{\p}{\mathcal{P}}
\newcommand{\teq}{\triangleq}

\title{Active Classification with Uncertainty Comparison Queries}

\author{
  Zhenghang Cui$^{\normalfont \text{1,2}}$\quad
  Issei Sato$^{\normalfont \text{1,2}}$\\
  $^\text{1}$ The University of Tokyo, Japan\\
  $^\text{2}$ RIKEN Center for Advanced Intelligence Project, Japan\\
  \texttt{\{cui, sato\}@g.ecc.u-tokyo.ac.jp}}

\begin{document}
\maketitle

\begin{abstract}
Noisy pairwise comparison feedback has been incorporated to improve the overall query complexity of interactively learning binary classifiers.
The \textit{positivity comparison oracle} is used to provide feedback on which is more likely to be positive given a pair of data points.
Because it is impossible to infer accurate labels using this oracle alone \textit{without knowing the classification threshold}, existing methods still rely on the traditional \textit{explicit labeling oracle}, which directly answers the label given a data point.
Existing methods conduct sorting on all data points and use explicit labeling oracle to find the classification threshold.
The current methods, however, have two drawbacks: (1) they needs unnecessary sorting for label inference; (2) quick sort is naively adapted to noisy feedback and negatively affects practical performance.
In order to avoid this inefficiency and acquire information of the classification threshold, we propose a new pairwise comparison oracle concerning uncertainties.
This oracle receives two data points as input and answers which one has higher uncertainty.
We then propose an efficient adaptive labeling algorithm using the proposed oracle and the positivity comparison oracle.
In addition, we also address the situation where the labeling budget is insufficient compared to the dataset size,
which can be dealt with by plugging the proposed algorithm into an active learning algorithm.
Furthermore, we confirm the feasibility of the proposed oracle and the performance of the proposed algorithm theoretically and empirically.
\end{abstract}

\section{Introduction}
Active learning studies interactive algorithms that can achieve the same generalization ability as passive learning, with more efficient query complexity.
It is known that active learning can achieve exponential improvement over passive learning under certain conditions \cite{hanneke14}.
However, this improvement does not always hold for more general situations.
Consequently, active learning methods have been developed by casting assumptions on the underlying data distribution and the target concept, or designing different forms of oracles that can better take advantage of the knowledge of annotators.

This paper focuses on the latter approach, specifically on methods incorporating the \textit{positivity comparison oracle} into active learning.
This oracle has high practicality in applications, as it has already been extensively used in other fields of machine learning, such as preference learning \cite{chu05, furnkranz10}.
It is obvious that using feedback from only this oracle, we can at most sort all unlabeled data points according to their positivity, or class-posterior probabilities.
Without knowing the location of the classification threshold to some extent, we cannot infer labels with accuracy guarantees.
Therefore, existing methods \cite{kane17, xu17} still need to access the \textit{explicit labeling oracle} to infer labels.

Among the existing methods, the one by Kane et al. \cite{kane17} takes a geometry approach, thus results a dimension-dependent query complexity.
Moreover, it only considers the noise-free setting of oracles, which limits its practicality.
Thus, we focus on the other method by Xu et al. \cite{xu17}, which shares a similar problem setting with this paper.
For $n$ unlabeled data points, the main idea of Xu et al. \cite{xu17} is (1) conducting quick sort using $\mathcal{O}(n\log{n})$ queries to the positivity comparison oracle and treat the feedback as if it is noise-free; (2) conducting binary search to locate the classification threshold using $\mathcal{O}(\log{n})$ queries to the explicit labeling oracle.
We note that knowing the positivity comparison order of all data points is unnecessary for the goal of label inference.
Given an unlabeled data point $x$, we are only interested in the relationship between $p(y=+1|x)$ and the classification threshold $0.5$, not the relationship between $p(y=+1|x)$ and class-posterior probabilities of other data points.
On the other hand, knowing all class labels cannot reconstruct the sorted list of class-posterior probabilities.
We recognize sorting over at least a subset of data points is inevitable due to collaborating with the explicit labeling oracle and the lack of information of the classification threshold.
Although it is feasible to improve the naive quick sort approach of the existing method, we choose to pose a question at a higher level:

\textit{``What form of feedback can efficiently provide (indirect) information of the classification threshold for the positivity comparison oracle?''}.

With this desired form of feedback, we would be able to save positivity comparison queries from unnecessary sorting.
Note this question is fundamentally motivated by the  Vapnik's principle \cite{vapnik98}.
\begin{quote}
If you possess a restricted amount of information for solving some problem, try to solve the problem directly and never solve a more general problem as an intermediate step.
\end{quote}
In this paper, our problem is the lack of information of the classification threshold.
To this end, we propose a new form of oracle, the \textit{uncertainty comparison oracle}, which asks annotators to compare the uncertainties of a pair of data points.
We assume that higher uncertainty indicates being closer to the classification threshold.
Properly using this new oracle, we can efficiently select the set of data points with high uncertainties, namely the set of data points that appears closet to the classification threshold.
Then, using this selected set as a delegation (a proxy or an approximation) of the unknown classification threshold, we can infer labels with accuracy guarantees to data points outside this selected set, which are the majority of unlabeled data points.
Not surprisingly, the explicit labeling oracle is no longer needed due to its inferior compatibility with pairwise comparisons.

In summary, for the problem of interactive labeling with access to the \textit{positivity comparison oracle}, our contributions are three-fold:
\begin{itemize}
 \item We propose a novel pairwise comparison oracle that compares the uncertainty of two unlabeled data points.
 \item We propose a feasible labeling algorithm effectively accessing the aforementioned two kinds of pairwise comparison oracles, as well as its applications under different query budgets.
 \item We establish the error rate bound for the proposed algorithm and generalization error bounds for its applications, and confirm their empirical performance.
\end{itemize}

\section{Labeling with pairwise comparisons}
In this section, we introduce the two comparison oracles and the proposed labeling algorithm.

\subsection{Preliminaries}
We consider the binary classification problem.
Let $\X\subset\R^d$ denote the $d$-dimensional sample space and
$\Y=\{+1,-1\}$ denote the binary label space.
Let $\p_{\X\Y}$ denote the underlying data distribution over $\X\times\Y$
and $\eta(x)\teq p(Y=+1|X=x)$ denote the underlying conditional probability for a data point $x$ being positive.
Then $h^*\teq\text{sign}(\eta(x)-0.5)$ is the Bayes classifier minimizing the classification risk $R(f)\teq\E_{(x,y)\sim\p_{\X\Y}}[\one_{f(x)\neq y}]$ for a classifier $f:\X\rightarrow\Y$.
In this problem setting, we are given only data points drawn from $\p_\X$, the marginal distribution over $\X$.
Without accessing class labels $\{h^*(x)|x\in\X\}$,
we query the following two oracles for essential information.

\subsection{Two pairwise comparison oracles}
\paragraph{Positivity comparison oracle}
This oracle receives two data points as input and
answers whether the first data point has a higher probability of being positive. The answer ``$+1$'' means ``yes" and ``$-1$'' means ``no''.
This is a common oracle that has been used in many different fields such as interactive classification \cite{kane17, xu17} and preference learning \cite{chu05, furnkranz10}.
We denote this oracle by $O_1:\X\times\X\rightarrow\{+1,-1\}$ and define it with the following noise condition.
\begin{condition}
\label{cond:pos}
Distribution $\p_{\X\Y}$ and oracle $O_1$ satisfies the condition with noise parameter $\epsilon_1\geq0$
if $\mathbb{E}_{x_1,x_2\sim\p_\X}[\mathbbm{1}_{O_1(x_1, x_2)(\eta(x_1) - \eta(x_2))<0}]=\epsilon_1$.
\end{condition}

\paragraph{Uncertainty comparison oracle}
This is our proposed oracle that receives two data points as input and
answers whether the first one has higher uncertainty.
The answer ``$+1$'' means ``yes'' and ``$-1$'' means ``no''.
We define the uncertainty of a data point $x\in\X$ as the difference between $\eta(x)$ and the classification threshold $0.5$.
This difference $|\eta(x) - 0.5|$ being small means $x$ has high uncertainty.
We denote this oracle by $O_2:\X\times\X\rightarrow\{+1,-1\}$ and define it with the following noise condition.

\begin{condition}
\label{cond:uncer}
Distribution $\p_{\X\Y}$ satisfies this condition with noise parameter $\epsilon_2\geq0$
if $\mathbb{E}_{x_1,x_2\sim\p_\X}[\mathbbm{1}_{O_2(x_1, x_2)(|\eta(x_2)-0.5|-|\eta(x_1)-0.5|)<0}]=\epsilon_2$.
\end{condition}

Note that the above conditions only assume the error rates.
Thus, answers may not hold for a proper order.
Namely, it is possible to collect positive answers from $O_1(x_1, x_2)$ and $O_1(x_2, x_1)$ (asymmetricity), or $O_1(x_1, x_2)$, $O_1(x_2,x_3)$ and $O_1(x_3,x_1)$ (intransitivity) for $x_1,x_2,x_3\sim\p_\X$.
The same holds for $O_2$.
Therefore our assumptions are relatively weak compared to parametric models, such as the Bradley-Terry-Luce (BTL) model \cite{bradley, luce}.

\subsection{Proposed labeling algorithm}
We propose a labeling algorithm that does not access the explicit labeling oracle at all and can still infer accurate labels.
Given a set of unlabeled data points $D$ sampled from $\p_\X$ with size $n$,
the idea is to first select a subset of $t$ data points $D'\subset D$ as a proxy or delegation for the classification threshold where $t\ll n$.
Note that we do not need to know the ranking order based on class-posterior probabilities of either $D'$ or $D\setminus D'$, and we want to find this subset by actively accessing the oracle as few times as possible.
This can be formulated as a top-$t$ items selection problem from noisy comparisons and has been extensively studied.
Note that we want to select the most uncertain data points, thus the uncertainty comparison oracle $O_2$ will be queried.
To this end, we choose the theoretical-guaranteed and practically promising algorithm proposed by Mohajer et al. \cite{mohajer17} as the first step of our algorithm.
For the self-containment of this paper, we briefly introduce this algorithm and its theoretical property.

\paragraph{Top selection algorithm from noisy comparisons \cite{mohajer17}}
Suppose we want to select $D'$ of $t$ data points from $D$ of $n$ data points based on a noisy comparison oracle.
The algorithm can be described in following steps.
\begin{enumerate}
  \item Separate the whole set into $t$ subsets with equal size of around $\frac{n}{t}$.
  \item On each subset, conduct a single-elimination tournament to select a single data point which is supposed to have highest uncertainty. Because the comparison results are noisy, each comparison is repeated $m$ times where $m$ is a hyper-parameter.
  \item For the $t$ data points selected from $t$ subsets, construct a heap structure.
  \item Move the data point at the top of the heap to $D'$.
  \item Conduct a single-elimination tournament with $m$ repetitions on the subset of which the first element of $D'$ belongs to.
  \item Reconstruct the heap and move the data point at the top of the heap to $D'$.
  \item Repeat the step five and six until $D'$ has enough data points.
\end{enumerate}

Although the above algorithm is a simple combination of single-elimination tournament and a heap structure, it enjoys the following favorable query complexity bound.
\begin{theorem}[Mohajer et al. \cite{mohajer17}]
With probability exceeding $1-(\log{n})^{-c_0}$, the subset of top-$t$ instances can be identified by the above algorithm with the query complexity upper bounded by $c_1(n+t\log{t})\frac{\max(\log\log{n}, \log{t})}{(\epsilon_2-0.5)^2}$.
Here, $c_0, c_1$ are some universal positive constants.
\end{theorem}

After selecting $t$ most uncertain data points using the above algorithm, we use this selected subset $D'$ and the positivity comparison oracle $O_1$ to decide labels of $D\setminus D'$.
The whole algorithm can be summarized in following three steps.
\begin{enumerate}
  \item Use $O_2$ to find $D'$, a subset of $t$ most uncertain data points.
  \item For each data point $x\in D\setminus D'$, we use $O_1$ to compare it with all (or part of) data points in $D'$ to infer its label by majority votes.
  \item Since we do not assume $D$ is i.i.d. sampled so that we can use this algorithm in more general situations, the worst case could happen for any labeling of $D'$. Therefore, we can randomly label $D'$, or repeat the whole algorithm using $D'$ as the initial input.
\end{enumerate}
This algorithm can efficiently infer labels without requiring unnecessary information such as the ranking order of class-posterior probabilities.
The algorithm is formally described in Algorithm \ref{alg:proposed}.
An error rate bound for inferred labels under noise conditions is established in Section \ref{thm-sec:proposed}.

\begin{algorithm}
 \caption{Proposed Labeling Algorithm}
 \label{alg:proposed}
 \begin{algorithmic}[1]
  \REQUIRE Positive integer $t$, dataset $D$ with size $n$.
  \STATE Select $t$ most uncertain data points from $D$ using the algorithm of Mohajer et al. \cite{mohajer17} and $O_2$. Denote the selected set as $D'$.
  \FOR{$x_i \in D \setminus D'$}
   \STATE \textbf{If} $\sum_{x_j \in D'} O_1(x_i, x_j) \geq \frac12$,
          \textbf{then} let $\hat{y_i} \leftarrow 1$,
          \textbf{else} let $\hat{y_i} \leftarrow 0$.
  \ENDFOR
  \STATE Randomly label $x_i \in D'$.
  \ENSURE Inferred labels $\hat{Y}\teq\{\hat{y_i}\}_{i=1}^n$.
 \end{algorithmic}
\end{algorithm}

\subsection{Learning classifiers under different budgets}
For down-stream tasks,
we can feed $D$ and $\hat{Y}$ into any algorithms that rely on samples from $\p_{\X\Y}$.
In this paper,
we consider the general application of learning a binary classifier.

\paragraph{Sufficient budget case}
In this case, we assume enough budget for running Algorithm \ref{alg:proposed} on the whole dataset.
Then, we can obtain the inferred labels and feed them into any classification algorithms.
In this paper,
we consider the simplest non-parametric $k$-NN algorithm \cite{knn},
which is easy to implement and enjoys good theoretical guarantees.
A generalization bound for classifiers obtained in this case is established in Section \ref{thm-sec:knn}.

\paragraph{Insufficient budget case}
In this case, we consider a more practical situation
where the dataset is too large compared to the budget; thus, we cannot afford to run Algorithm \ref{alg:proposed} on the whole dataset.
We resort to using active learning with Algorithm \ref{alg:proposed} as a subroutine for the selected batch at each step.
The same as Algorithm 3 of Xu et al. \cite{xu17}, we consider a disagreement-based active learning algorithm calling the proposed Algorithm \ref{alg:proposed} at each step.
Algorithm \ref{alg:a2} describes the detailed algorithm.
\begin{algorithm}
 \caption{Disagreement-based active learning algorithm (Algorithm 3 of Xu et al. \cite{xu17}).}
 \label{alg:a2}
 \begin{algorithmic}[1]
  \REQUIRE $\epsilon$, a sequence of $n_i$, hypothesis set $H$.
  \STATE $H_1\leftarrow H$
  \FOR{$i=1,2,\cdots,\ceil{\log(\frac1\epsilon)}$}
   \STATE $S_i\leftarrow$ i.i.d. sample from $\p_\X$ with size $n_i$.
   \STATE $D_i\leftarrow\text{DIS}(S_i, H_i)$.
   \STATE Run Algorithm \ref{alg:proposed} with $\epsilon_i=\frac1{2^{i+2}}$ and $D_i$, obtain $\{\hat{y_j}\}_{j=1}^{|D_i|}$.
   \STATE $H_{i+1} \leftarrow \{h\in H_i: \sum_{j=1}^{n_i} \mathbbm{1}_{h(x_j)\neq\hat{y}_j} \leq \epsilon_i n_i\}$
  \ENDFOR
  \ENSURE Any Classifier in $H_{i+1}$
 \end{algorithmic}
\end{algorithm}

\section{Theoretical analysis}
In this section, we establish the error rate bound for Algorithm \ref{alg:proposed} and generalization error bounds for the $k$-NN algorithm and Algorithm \ref{alg:a2}.

\subsection{Analysis of the proposed labeling algorithm}
\label{thm-sec:proposed}
\begin{theorem}[Error rate bound]
\label{thm:proposed}
Suppose Condition \ref{cond:pos} and Condition \ref{cond:uncer} hold for $\epsilon_1, \epsilon_2\in[0,0.5)$.
Let $t = \Omega\left(\frac{\log2}{2(0.5-\epsilon_1)^2}\right)$.
Fix $\epsilon > 0$ and assume $D$ to be a set of size $n > \frac{t}{\epsilon}$
that contains data points $x\in\mathcal{X}$.
Then, there exist constants $C_1$ and $C_2$ such that for an execution of Algorithm \ref{alg:proposed} on $D$
with parameters $t$ and
$m \geq \frac{C_1\max(\log\log{n}, \log{t})}{(0.5-\epsilon_2)^2}$,
with probability at least $1-\delta$
where we denote $\delta\triangleq\delta(C_2, n, t, \epsilon_1)$ for simplicity,
the error rate of inferred labels is bounded as $\frac{|\{i\in[n]|\hat{y_i} \neq h^*(x_i)\}|}n \leq \epsilon$.
The query complexity is $\mathcal{O}\left(\frac{n}{\epsilon_1^2}\right)$ for $O_1$
and $\mathcal{O}\left(\frac{n\log\log{n}}{\epsilon_2^2}\right)$ for $O_2$.
\end{theorem}
Proof can be found in Appendix \ref{supp:proposed}.
Note that there are two hyper-parameters for the algorithm: the size of the delegation subset $t$ and the repetition number for each comparison $m$.
The above theory shows a principled way to select the hyper-parameter $t$,
which only depends on the error condition $\epsilon_1$.
For a reasonable range of $\epsilon_1\leq0.4$,
Algorithm \ref{alg:proposed} only requires $t$ to be at most $35$,
which is relatively small compared to the size of a modern dataset.
For the other hyper-parameter $m$, we empirically observe that a surprisingly small value, even $1$, shows promising performance.
For the query complexities, the $\mathcal{O}(n)$ factor should be required by at least one oracle, since we cannot decide the label of a data point without accessing it at least once.

\subsection{Analysis of nearest neighbors classifiers}
\label{thm-sec:knn}
We establish a generalization error bound for classifiers obtained by combining Algorithm \ref{alg:proposed} and $k$-NN.
We want to estimate the function $\eta(x)$ from the inferred labels by Algorithm \ref{alg:proposed}.
For $x\in \mathcal{X}$,
we denote indices of other points in a descending distance order by $\{\tau_q(x)\}_{q=1}^{n-1}$.
This means that for a metric $\rho$, it holds $\rho(x, x_{\tau_q}) \leq \rho(x, x_{\tau_{q+1}})$ for $q\in[1, n-2]$.
Thus, we can denote the resulting $k$-NN classifier as $\hat{f}(x;k) = \frac1k\sum_{q=1}^k \hat{y}_{\tau_q(x)}$.

We then introduce two essential assumptions.
First, we need a general assumption for achieving fast convergence rates for $k$-NN classifiers.
\begin{assumption}[Measure smoothness \cite{chaudhuri14}]
\label{asm:smooth}
With $\lambda>0$ and $\omega>0$, for all $x_1, x_2\in\mathcal{X}$, it satisfies
$$|\eta(x_1)-\eta(x_2)| \leq \omega\mu\left(B_{\rho(x_1, x_2)}(x_0)\right)^\lambda,$$
where $B_{\rho(x_1, x_2)}(x_0)$ denotes a ball with center $x_0$ and radius $\rho(x_1, x_2)$.
\end{assumption}

Then, we need the following Tsybakov's margin condition \cite{mammen99},
which is a common assumption for establishing fast convergence rates.
\begin{assumption}[Tsybakov's margin condition]
\label{asm:tsyba}
There exist $\alpha\geq0$ and $C_\alpha\geq1$ such that for all $\xi>0$ we have
\begin{equation*}
 \mu\left(\bigg\{x\in\mathcal{X}:0<\left|\eta(x)-\frac12\right|<\xi\bigg\}\right)\leq C_\alpha\xi^\alpha.
\end{equation*}
\end{assumption}

Finally, we establish the generalization error bound.
\begin{theorem}[Generalization error bound for $k$-NN]
\label{thm:knn}
Suppose the conditions for Theorem \ref{thm:proposed} hold.
Let the input and the output of Algorithm \ref{alg:proposed} be $D=\{{x_i}\}_{i=1}^n$ and $\hat{Y}=\{\hat{y}_i\}_{i=1}^n$.
Let $\hat{f}(x;k)$ be the $k$-NN classifier obtained and
$f^*(x) \triangleq \mathbbm{1}_{\eta(x)\geq\frac12}$ be the Bayes classifier.
Then, using the same notations as Theorem \ref{thm:proposed},
supposing that Assumption \ref{asm:smooth} holds with $\lambda>0$ and $\omega>0$, and
Assumption \ref{asm:tsyba} holds with $\alpha\geq0$ and $C_\alpha\geq1$,
for $\delta'\in(0,1)$, $4\log(\frac1{\delta'}) + 1\leq k\leq\frac{n}2$,
with probability at least $(1-\delta)(1-\delta')$,
we have $$R(\hat{f}) \leq R(f^*) + C_\alpha\left(\frac{2\epsilon}k + \omega\left(\frac{2k}{n}\right)^\lambda\right)^{\alpha+1}.$$
\end{theorem}
Proof can be found in Appendix \ref{supp:knn}.
The difference between the above bound and other generalization bounds under unknown asymmetric noise \cite{gao16, mohajer17} is that Theorem \ref{thm:knn} does not require the labels to be an i.i.d. sample from an underlying distribution, as they are instead inferred by Algorithm \ref{alg:proposed}.

\subsection{Analysis of disagreement-based active learning}
We establish the generalization error bound by the following corollary to justify Algorithm \ref{alg:a2}.
Its proof can be found in Appendix \ref{supp:a2}.

\begin{corollary}[Generalization error bound for active learning]
\label{thm:a2}
Suppose conditions for Theorem \ref{thm:proposed} hold.
Then, for an execution of Algorithm \ref{alg:a2}
with $\epsilon\in(0,1), \epsilon_i=\frac1{2^{i+2}}$,
with probability at least $1-\delta$, the output $\hat{h}$ satisfies $\text{P}_{x\sim\p_\X}[\hat{h}(x)\neq h^*(x)]\leq \epsilon$.
\end{corollary}

\section{Related work}
\paragraph{Weakly-supervised learning}
Learning classifiers from passively obtained comparisons without explicit class labels have been studied,
such as learning from similarity comparisons \cite{bao, shimada} and learning from triplet comparisons \cite{cui}.
However, the feasibility of learning in such cases relies on various inevitable assumptions. Bao et al. \cite{bao} assumes the group with more data to be the positive class.
The other two methods \cite{shimada, cui} assume specific data generation processes, which may not always hold in some applications.
Moreover, none of these methods have theoretical guarantees for noisy comparisons.
On the other hand, learning from totally unlabeled data has also been studied \cite{du, lu}. However, these methods require at least two datasets with different class priors $p(Y=+1)$, and they also need to know these class priors \textit{exactly}, which can be impossible in some cases.
The proposed labeling algorithm is transductive and can be combined with non-parametric classifiers, while above existing methods mainly rewrite the classification risk and require a differentiable model.
Furthermore, the proposed algorithm does not require above assumptions and additional information such as exact class priors.

\paragraph{Preference learning}
Results of $O_1$ are mainly used in this learning paradigm to learn a (partial) ranking over data points.
However, ranking cannot induce labels without further information as the class prior is needed to decide the classification threshold.
At the same time, labeling cannot induce ranking as there is no information on the ranking order of data with the same label.
Similar arguments also hold for bipartite ranking \cite{narasimhan13}.

\paragraph{Active learning}
Interactive classification with oracles that do not answer the explicit class labels has been studied \cite{beygelzimer16, kane17, xu17, tosh18}.
Beygelzimer et al. \cite{beygelzimer16} uses a search oracle that receives a function set as input and outputs a data point \textit{with its explicit class label}. Other two methods use the same oracle as $O_1$.
However, they all need to access the explicit labeling oracle.
On the other hand, Balcan et al. \cite{balcan12} uses only the class conditional queries (CCQ) without accessing the explicit labeling oracle.
However, labels can be inferred from a single CCQ query.
Although we cannot directly compare,
we claim that $O_2$ is weaker than CCQ because labels cannot be inferred from the query results of $O_2$.

\section{Experiments}
In this section, we confirmed the feasibility and performance of the proposed algorithm using both simulation and crowdsourcing data.

\subsection{Simulation study}
All experiments were repeated ten times on a server with an Intel(R) Xeon(R) CPU E5-2698 v4 @ 2.20GHz CPU and a Tesla V100 GPU.
Their mean values and standard deviations are reported.

\subsubsection{Sufficient budget case}
When considering constructing binary datasets from multi-class datasets, experiments in existing work usually split the whole dataset into two parts,
such as separating odd numbers and even numbers for hand written digits.
However, as the focus of the proposed oracle is the uncertainty, it is important to simulate experiments that has some kind of uncertainties in its expression.
For image datasets, the uncertainty can be expressed as visual similarity between two classes.
Therefore, we constructed following eight binary classification datasets that have visual similarity to some extent.
\begin{itemize}
 \item {\bf MNIST-a} denotes the MNIST \cite{mnist} data with label $1$ ($7877$ data) and $7$ ($7293$ data),
 \item {\bf MNIST-b} denotes the MNIST data with label $3$ ($7141$ data) and $5$ ($6313$ data).
 \item {\bf FMNIST-a} denotes the Fashion-MNIST \cite{fashion} data with label T-shirt/top ($7000$ data) and shirt ($7000$ data).
 \item {\bf FMNIST-b} denotes the Fashion-MNIST data with label pullover ($7000$ data) and coat ($7000$ data).
 \item {\bf KMNIST-a} denotes the Kuzushiji-MNIST \cite{kuzushi} data with the second label ($7000$ data) and eighth label ($7000$ data).
 \item {\bf KMNIST-b} denotes the Kuzushiji-MNIST data with the third label ($7000$ data) and seventh label ($7000$ data)
 \item {\bf CIFAR10-a} denotes the CIFAR-10 \cite{cifar} data with label automobile ($5000$ data) and truck ($5000$ data).
 \item {\bf CIFAR10-b} denotes the CIFAR-10 data with label deer ($5000$ data) and horse ($5000$ data).
\end{itemize}

For all datasets except CIFAR-10, a logistic regression classifier was first learned on all selected data with one hundred thousand maximum iteration.
The oracles were then simulated using the output conditional probabilities of this logistic regression classifier.
For CIFAR-10, a ResNet152 \cite{resnet} classifier was first trained on the whole dataset ($10$ classes) for $100$ epochs.
Then, the $2048$-dimension expressions before the last fully connected layer were used as low dimensional features, which were then used to train a logistic regression classifier.
The logistic regression classifier and the $k$-NN classifiers are trained on these $2048$-dimension features instead of the original input.
We set $k=5$ for $k$-NN classifiers throughout all experiments.
We randomly split training and test set according to the $4:1$ ratio for every repetition of the algorithms.
We do not have sensitive hyper-parameters to tune, thus we did not separated a validation set.
For the Co-teaching experiments, we set batchsize as $1024$ and epochs as $100$. We adopted the public codes provided by the authors, thus followed all other default settings therein, such as learning rate schedule.

We considered the conservative case where the noise rates are high and the repetition number $m$ is small.
Theorem \ref{thm:proposed} indicates that the size of the delegation subset $t$ usually has a maximum of $35$, thus we set $t$ to be $10$ or $35$.
Table \ref{t1} shows that a larger set of delegation set (corresponding to a higher $t$) contributes to a better label accuracy, thus a better generalization ability.
This behavior matches the expectation as the inferred label for each non-delegation data point becomes more accurate.
We also observed that even with a small $t$, $k$-NN shows promising generalization ability.

\begin{table*}[htbp]
\centering
\caption{Performance when the repetition number $m=1$, noise rates $\epsilon_1=\epsilon_2=0.4$.}
\label{t1}
\begin{tabular} {ccccccc}
\hline
\text{Dataset} & \thead{Label\\Accuracy\\($t$=10)} & \thead{$k$-NN\\Test Accuracy\\($t$=10)} & \thead{Label\\Accuracy\\($t$=35)} & \thead{$k$-NN\\Test Accuracy\\($t$=35)}\\\hline
\text{MNIST-a} & 67.89 (0.37) & 77.63 (0.83) & 80.94 (0.47) & 92.36 (0.60)\\
\text{MNIST-b} & 67.10 (0.52) & 76.11 (0.79) & 80.46 (0.37) & 92.93 (0.37)\\
\text{FMNIST-a} & 65.78 (0.26) & 70.96 (0.45) & 76.38 (0.20) & 81.40 (0.19)\\
\text{FMNIST-b} & 66.25 (0.34) & 72.28 (0.50) & 77.25 (0.24) & 83.36 (0.20)\\
\text{KMNIST-a} & 68.69 (0.56) & 78.90 (1.07) & 81.64 (0.62) & 94.30 (0.58)\\
\text{KMNIST-b} & 67.99 (0.26) & 77.45 (0.45) & 78.88 (0.36) & 90.16 (0.33)\\
\text{CIFAR10-a} & 69.34 (0.44) & 80.09 (0.82) & 82.07 (0.41) & 94.28 (0.31)\\
\text{CIFAR10-b} & 68.67 (0.20) & 78.47 (0.59) & 81.83 (0.50) & 93.95 (0.42)\\
\hline
\end{tabular}
\end{table*}

Table \ref{t2} shows the results of the optimism situation when the noise rates were low and sufficient budget for a larger $m$ was available.

\begin{table*}[htbp]
\centering
\caption{Performance when repetition $m=10$, noise rate $\epsilon_1=\epsilon_2=0.1$.}
\label{t2}
\begin{tabular} {ccccccc}
\hline
\text{Dataset} & \thead{Label\\Accuracy\\($t$=10)} & \thead{$k$-NN\\Test Accuracy\\($t$=10)} & \thead{Label\\Accuracy\\($t$=35)} & \thead{$k$-NN\\Test Accuracy\\($t$=35)}\\\hline
\text{MNIST-a} & 99.74 (0.01) & 99.39 (0.03) & 99.84 (0.01) & 99.35 (0.03)\\
\text{MNIST-b} & 97.12 (0.03) & 98.36 (0.09) & 97.22 (0.02) & 98.36 (0.06)\\
\text{FMNIST-a} & 87.19 (0.06) & 83.95 (0.18) & 87.38 (0.06) & 84.14 (0.16)\\
\text{FMNIST-b} & 88.84 (0.04) & 86.26 (0.20) & 88.86 (0.04) & 86.67 (0.18)\\
\text{KMNIST-a} & 98.78 (0.01) & 99.12 (0.05) & 98.90 (0.01) & 99.00 (0.02)\\
\text{KMNIST-b} & 92.33 (0.03) & 94.53 (0.14) & 92.36 (0.03) & 94.85 (0.09)\\
\text{CIFAR10-a} & 99.87 (0.02) & 99.92 (0.02) & 99.97 (0.01) & 99.95 (0.01)\\
\text{CIFAR10-b} & 99.86 (0.01) & 99.98 (0.01) & 99.94 (0.01) & 99.98 (0.01)\\
\hline
\end{tabular}
\end{table*}

We next confirmed the quality of inferred labels using a more powerful model.
Co-teaching \cite{bohan} is a recently proposed training method for extremely \textit{noisy labels}.
It holds two classifiers which feed their small loss data points to the other classifier for training.
Although lacking theoretically guarantees,
it is reported promising performance \cite{bohan}.
We used relatively small ResNet18 \cite{resnet} models and restrain from tuning any hyper-parameters for Co-teaching.

\begin{figure}[htbp]
  \centering
  \subfloat{
    \includegraphics[width=0.46\textwidth]{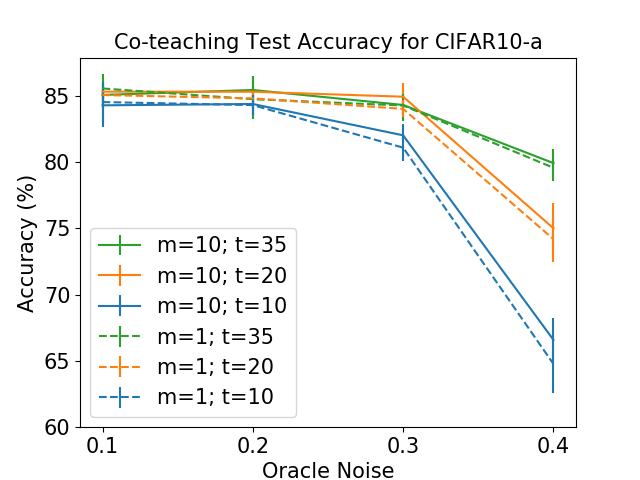}}
  \subfloat{
    \includegraphics[width=0.46\textwidth]{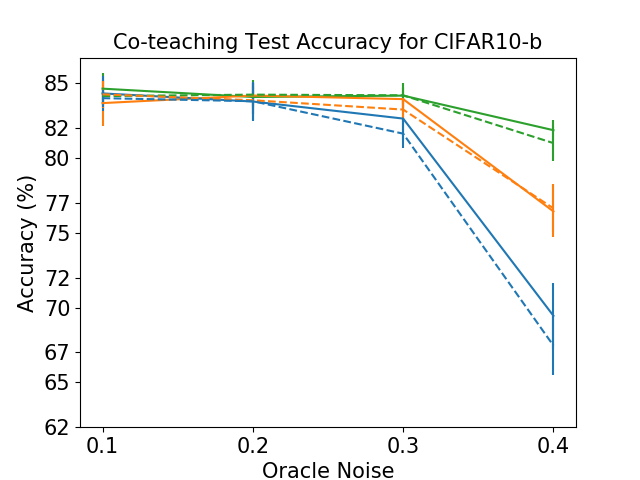}}
  \caption{Generalization performance of co-teaching classifiers.}
  \label{fig:cifar}
\end{figure}

Figure \ref{fig:cifar} shows results with same size of delegation set in the same color, and uses dot lines to show results with fewer repetition numbers.
We observe that setting $m=1$ already shows promising accuracy, with $t$ set to be the theoretical maximum $35$.
For the same value of $t$,
increasing $m$ from $1$ to $10$ can offer only little improvement on the accuracy.
Setting $m$ to $1$ means we only query each pair once and proceed the algorithm believing the answer is correct.
This shows that the proposed algorithm is highly robust to query noise, as it shows promising performance using the single noisy result without repeating the same query many times.
Moreover, the low noise rate regime shows comparable performance under different settings, which means the proposed algorithm can generally achieve high performance with low budget.

\begin{figure}[htbp]
  \centering
  \subfloat{
    \includegraphics[width=0.48\textwidth]{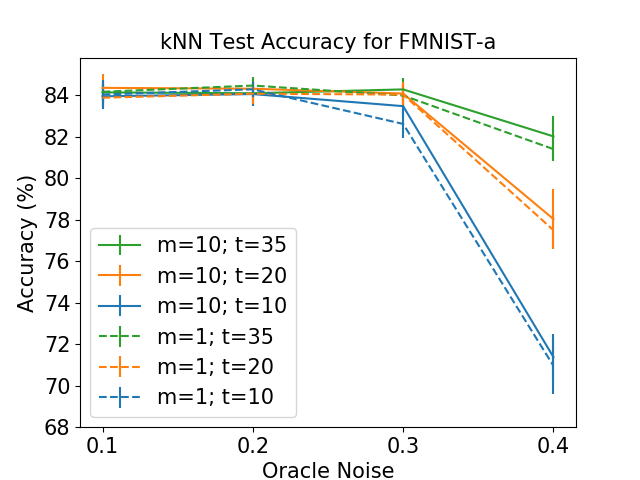}}
  \subfloat{
    \includegraphics[width=0.48\textwidth]{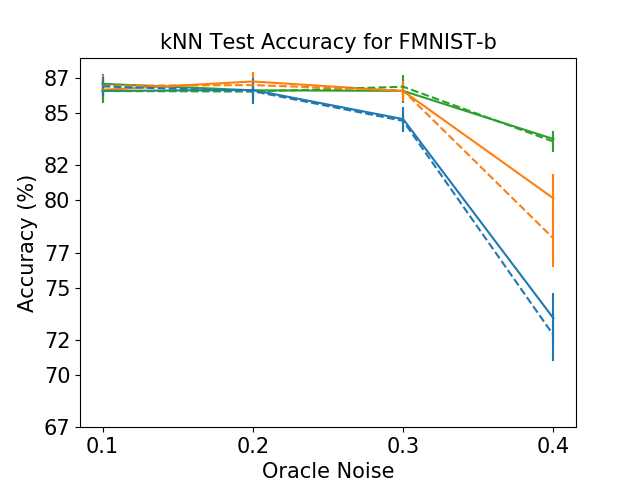}}
  \caption{Generalization performance of $k$-NN classifiers for Fashion-MNIST datasets.}
  \label{fig:fmnist}
\end{figure}

Figure \ref{fig:fmnist} shows the detailed investigation on the Fashion MNIST datasets. It shows similar tendency as the previous Co-teaching results on CIFAR-10 datasets.

\subsubsection{Insufficient budget case}
In this case, because Algorithm \ref{alg:a2} needs to loop over every available hypothesis at each step, it is infeasible to start with a large hypotheses set.
Note that even for the MNIST dataset with $784$ features and the simplest linear models, using a discrete exploring space of size $10$ for the parameter corresponding to each feature creates a huge hypotheses set of size $10^{784}$.
Therefore, in order to illustrate the feasibility of the algorithm, we used $2$-dimensional toy data generated from two Gaussian distributions that are symmetric to the origin point.
Specifically, we used two Gaussian distributions with mean value of $(2, 2)$ and $(-2, -2)$ and the identical matrix as both covariances.
From these distributions, we drew ten thousand data points in total, with each data point having an equal probability to be generated from either distribution.
Then a logistic regression classifier is trained with one hundred thousand maximum iteration to simulate the oracles.
For the hypothesis set, we used $1000$ equally separated linear classifiers passing through the origin point.
Setting the desiring precision $\epsilon=0.1$ resulted three steps based on Algorithm \ref{alg:a2}.
Table \ref{table:active} shows the number of left candidate hypotheses and their test accuracy at each step.

\begin{table*}[htbp]
\centering
\caption{Active Learning Experiment Results}
\label{table:active}
\begin{tabular} {cccc}
\hline
\text{} & Step 1 & Step 2 & Step 3\\ \hline
\thead{Number of\\Left Hypotheses} & 674.10 (4.97) & 525.60 (7.34) & 196.90 (71.85)\\
\hline
\thead{Test Accuracy of\\Left Hypotheses} & 96.98 (0.44) & 99.29 (0.19) & 99.78 (0.11)\\
\hline
\end{tabular}
\end{table*}

\subsection{User study}
The previous section investigated the proposed algorithm using artificial oracles, and the feasibility in real-world situations remains untouched.
Therefore, we conducted user study using crowdsourcing in this section.

\subsubsection{Character recognition task}
\label{sec:userexp1}
In this task, we focused on the classification of \textit{Kuzushiji} (cursive Japanese) \cite{kuzushi}, which is important for advocating research on Japanese historical books and documents.

\paragraph{Goals}
We want to justify the proposed oracle and confirm whether the proposed algorithm can work on results collected through user study \textit{without simulation}.
Specifically, we want to
(1) confirm whether data pairs selected by the proposed algorithm are easier for uncertainty comparison than explicit labeling, and
(2) confirm whether the proposed algorithm can work on only crowdsourcing results.
We will introduce the data and the general interface we used in user study, followed by detailed description of each user study setting in the following paragraphs.

\paragraph{Data}

\begin{wrapfigure}{r}{0.25\textwidth}
  \vspace{-25pt}
  \subfloat{
    \includegraphics[width=0.12\textwidth]{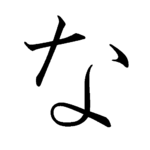}}
  \subfloat{
    \includegraphics[width=0.12\textwidth]{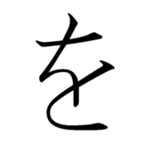}}
  \caption{Sample images for `NA' in the left and `WO' in the right.}
  \vspace{-10pt}
  \label{fig:nawo}
\end{wrapfigure}

From the Kuzushiji-MNIST dataset \cite{kuzushi},
we selected the $5$-th and the $10$-th characters to form the binary classification task.
The reading alphabet is `NA' for the $5$-th character and `WO' for the $10$-th character.
Figure \ref{fig:nawo} shows them in a standard font.
Albeit the visual similarity, these two characters are important auxiliary words with distinct meanings.
Thus, wrongly recognizing the two characters can harm the understanding of the sentence.
This recognition task has a natural affinity with uncertainty comparison, as in daily writing, the difficulty of recognizing a hand written character is easier to interpret, rather than recognizing the exact character.

\paragraph{Methods}
We prepared three types of questions:
explicit labeling,
pairwise positivity comparison,
and pairwise uncertainty comparison.
We also asked annotators for the difficulty of each question when necessary.
The interfaces are shown by the following list.

\begin{figure}[htbp]
\centering
\includegraphics[width=.7\textwidth]{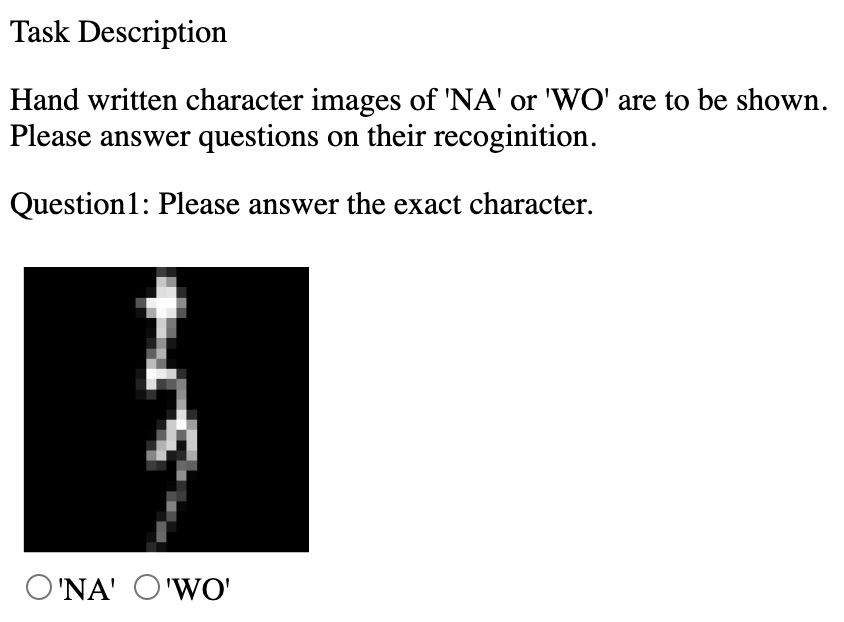}
\caption{Questionnaire of explicit labeling.}
\label{fig:absolute}
\end{figure}

\begin{figure}[htbp]
\centering
\includegraphics[width=.7\textwidth]{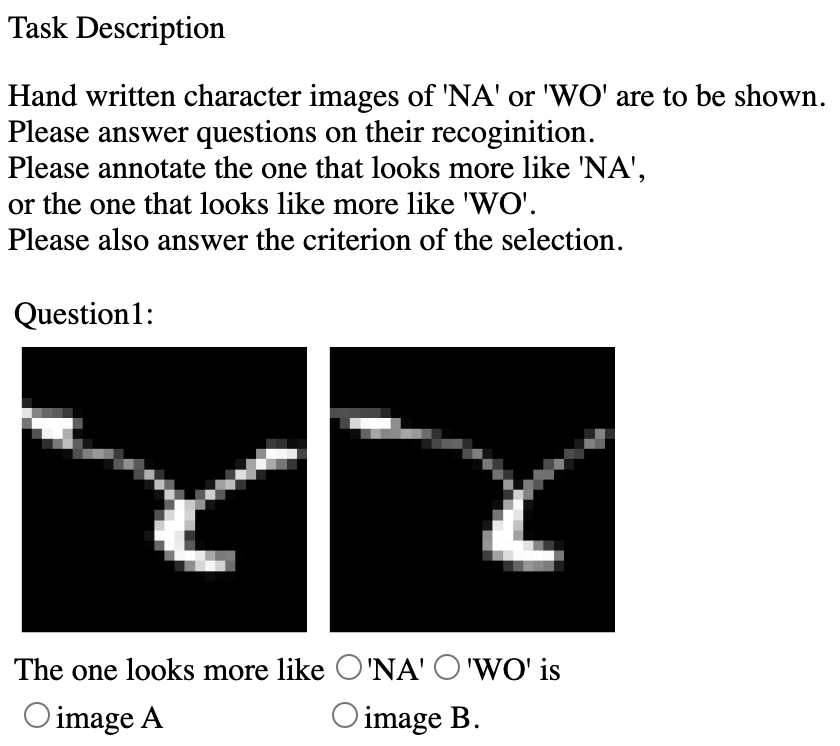}
\caption{Questionnaire of pairwise positivity comparisons.}
\label{fig:positivity}
\end{figure}

\begin{figure}[htbp]
\centering
\includegraphics[width=.8\textwidth]{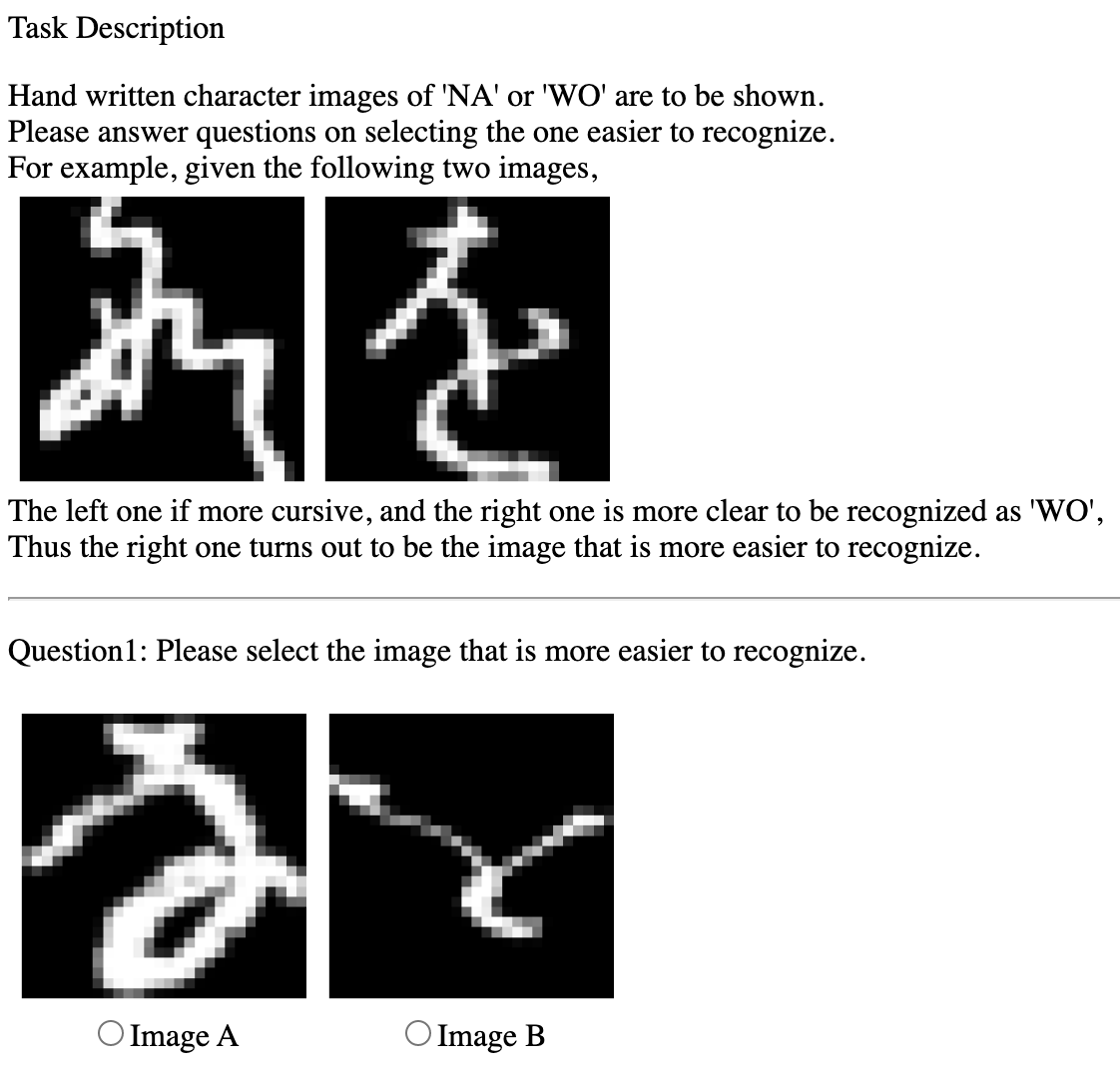}
\caption{Questionnaire of pairwise uncertainty comparisons.}
\label{fig:uncertainty}
\end{figure}

\begin{figure}[htbp]
\centering
\includegraphics[width=.8\textwidth]{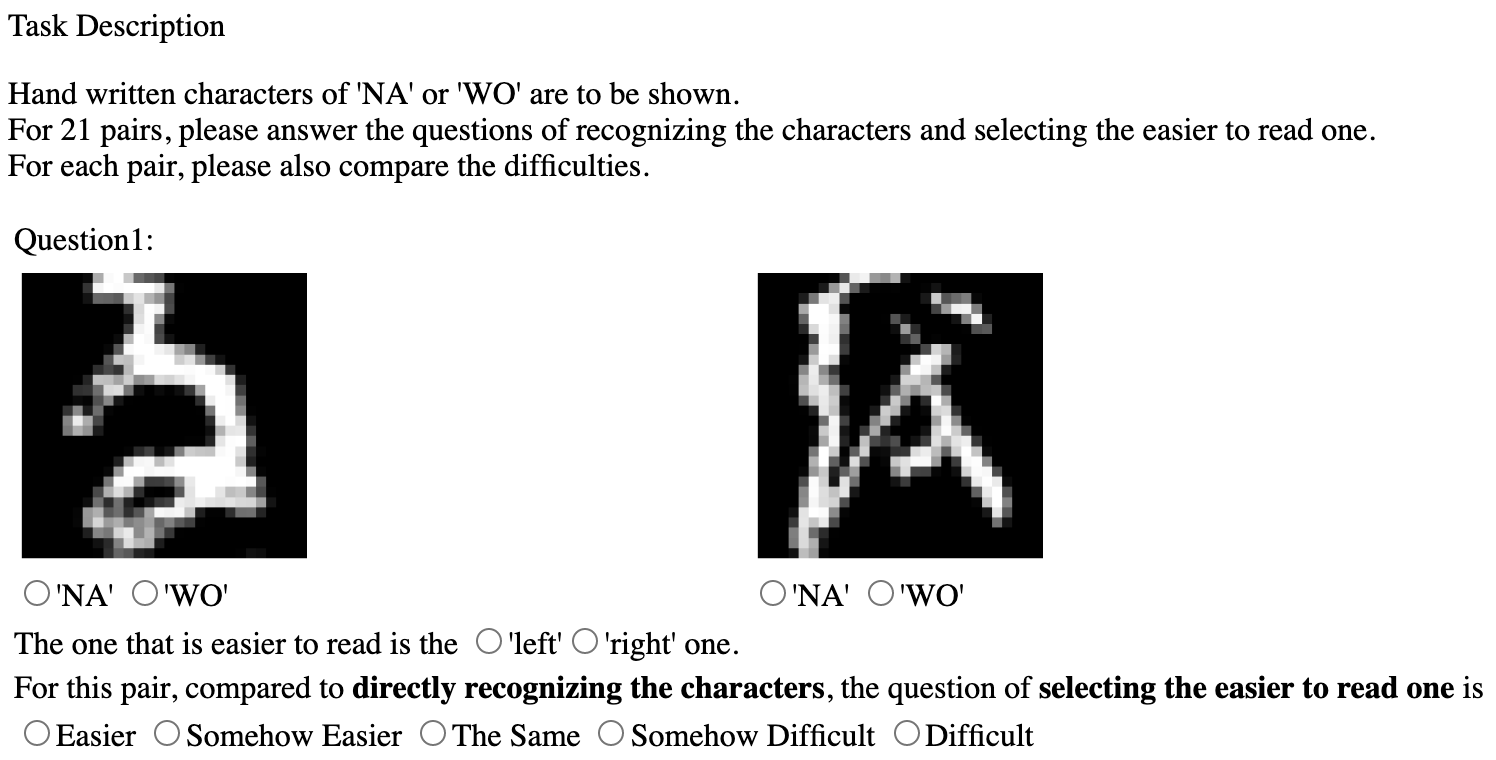}
\caption{Questionnaire of difficulty evaluation of pairwise uncertainty comparisons.}
\label{fig:pair-difficulty}
\end{figure}

\begin{itemize}
  \item Figure \ref{fig:absolute} shows how we ask annotators for explicit labels.
  \item Figure \ref{fig:positivity} shows how we ask annotators for pairwise positivity comparisons. If we fix one label such as `NA' and ask which one is more likely to be `NA', there are cases that both images in a pair look similar to `WO', thus it's difficult to answer. Therefore, we also ask annotators to choose either `NA' or `WO' that is used as the criterion of positivity.
  \item Figure \ref{fig:uncertainty} shows how we ask annotators for pairwise uncertainty comparisons. As this is a newly proposed comparison question and annotators may be not used to answer it, we give an explanatory example on how to select.
  \item Figure \ref{fig:pair-difficulty} shows how we ask annotators for difficulty evaluation of uncertainty comparisons compared to explicit labeling. We asked annotators to answer both queries first to familiarize them with the problem.
\end{itemize}

\paragraph{Justification for uncertainty comparisons}
In this user study, we confirmed whether the data pairs selected by the algorithm for $O_2$ are difficult for explicit labeling.
We first greedily selected $25$ medoids from all data points.
Then, we ran the proposed algorithm on these $25$ data points using artificial oracles, and collected the $42$ pairs that were selected for $O_2$.
Finally, we conducted user study from $50$ annotators on explicit labeling and uncertainty comparison on these $42$ pairs.
For each, we also asked the difficulty of uncertainty comparisons compared to explicit labeling using scores from one to five, with a \textit{smaller} score indicating an \textit{easier} question.
Furthermore, we collected difficulty evaluation of explicit labeling for each image from $10$ annotators.

In order to investigate the difficulty evaluation on pair attributes, we introduce the \textit{individual difficulty} for each single image. Another difficulty will be introduced in the following paragraph.
Then, based on the user evaluation of \textit{individual difficulties}, we classified data pairs into three types:
(1) the `E' type containing two easy data points,
(2) the `\&' type containing one easy and one difficult data point, and
(3) the `D' type containing two difficult data points.

We then aggregated the user evaluations based on pair types.
Table \ref{table:uncertainty-difficulty} shows the mean and standard deviation of the difficulty evaluations for each type, as well as t statistics and p values when conducted one sample t test against value $3$, which means two types of query have equal difficulty.
From the results, we can conclude that as pair type changes from `E' to `D', uncertainty comparison becomes less favored against explicit labeling.
For type `D', the mean of difficulty evaluations is not significantly different from $3$, as the p value $0.06>0.05$.
However, as the proposed algorithm focuses on separating difficult images, random decisions on images with similar difficulty do not harm the performance.

\begin{table*}[htbp]
\centering
\caption{Statistics of difficulty evaluation of uncertainty comparisons.}
\label{table:uncertainty-difficulty}
\begin{tabular} {ccccc}
\hline
\text{} & Type `E' & Type `\&' & Type `D' & Total \\ \hline
\thead{Number of Pairs} & 12 & 25 & 5 & 42 \\
\thead{Number of\\Total Evaluations} & 600 & 1250 & 250 & 2100 \\
\hline
\thead{Mean} & 2.57 & 2.82 & 2.84 & 2.75 \\
\thead{Standard Deviation} & 1.28 & 1.38 & 1.35 & 0.34 \\
\hline
\thead{t statistic} & -8.23 & -4.68 & -1.91 & -5.15 \\
\thead{p value} & $1.19\times10^{-15}$ & $3.22\times10^{-6}$ & 0.06 & $4.69\times10^{-6}$ \\
\hline
\end{tabular}
\end{table*}

\paragraph{Algorithm feasibility using simulated pairwise comparisons}
In this user study, we first greedily selected $50$ medoids as training data.
We then collected explicit label feedback from $20$ annotators.
For a single image in these $50$ medoids, we simulated its class probability by the proportion of class assignments in the $20$ evaluations.
For example, if $15$ annotators assigned positive label to an image, we defined its probability to be positive as $\frac{15}{20}=0.75$.
These probabilities were then used to simulate both kinds of pairwise comparisons.
Using inferred labels as input, the last layer of a pre-trained neural network was fine-tuned.

\begin{figure}[htbp]
  \centering
  \subfloat{
    \includegraphics[width=0.49\textwidth]{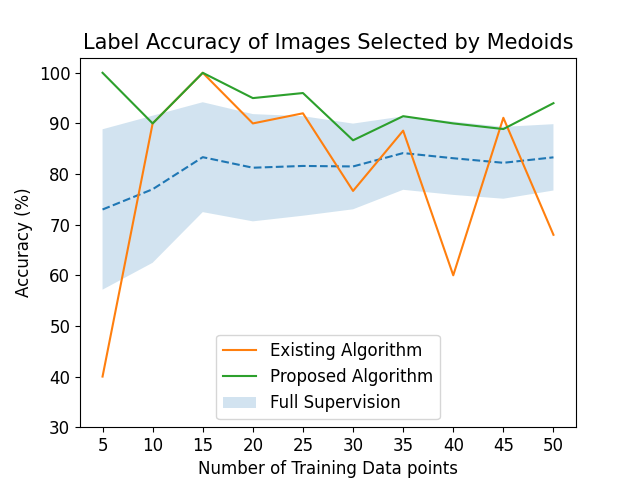}}
  \subfloat{
    \includegraphics[width=0.49\textwidth]{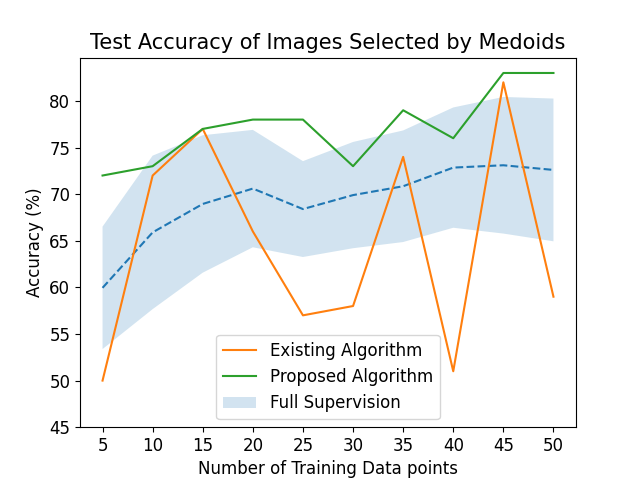}}
  \caption{Performance using simulated pairwise comparisons.}
  \label{fig:kmed-sim}
\end{figure}

Figure \ref{fig:kmed-sim} shows the label accuracy and the test accuracy when using different numbers of medoids as training data.
The test accuracy measures the performance of each classifier learnt from inferred labels on a test dataset of size $100$, which is uniformly selected without replacement excluding the training data points.
It can be clearly observed for the full supervision case that more training data contribute higher accuracies.
It is not clear for the other two methods, because they rely on not only the number of training data, but also the quality of their pairwise comparison feedback.
Although there were $64\%$ ties among all uncertainty pairwise comparisons, the proposed method showed consistent performance.
However, with $24\%$ ties among all positivity pairwise comparisons, the existing method failed to perform consistently, even with parameter tuning.

\paragraph{Algorithm feasibility using user feedback on pairwise comparisons}
In this user study, we confirmed the performance of each algorithm on only crowdsourcing results.
We greedily selected $25$ medoids \cite{kmed},
collected answers for all possible combinations among these medoids from $10$ annotators, and used aggregated majority as input to the existing algorithm \cite{xu17} using both positivity comparisons and explicit class labels and the proposed algorithm.
We adopted a pre-trained neural network and fine-tuned its last layer considering the small number of training data.

\begin{figure}[htbp]
  \centering
  \subfloat{
    \includegraphics[width=0.49\textwidth]{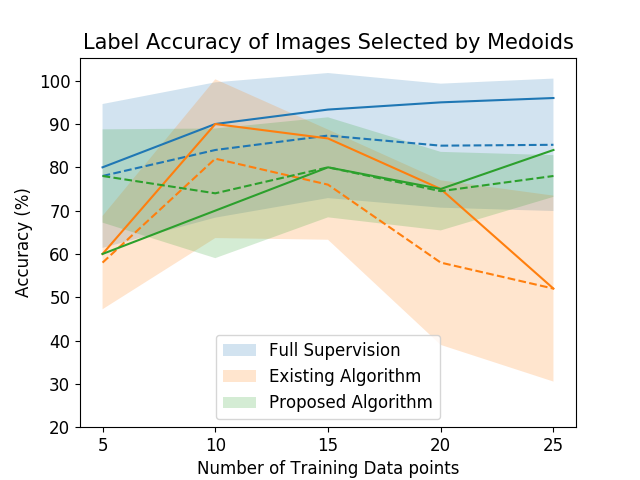}}
  \subfloat{
    \includegraphics[width=0.49\textwidth]{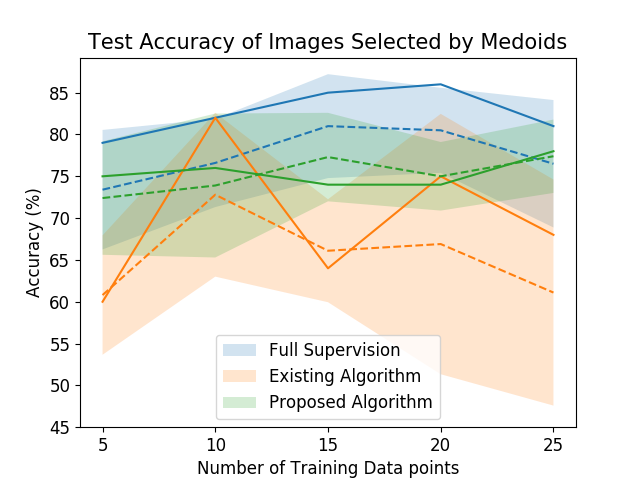}}
  \caption{Performance on medoids.}
  \label{fig:kmed}
\end{figure}

\begin{figure}[htbp]
  \centering
  \subfloat{
    \includegraphics[width=0.49\textwidth]{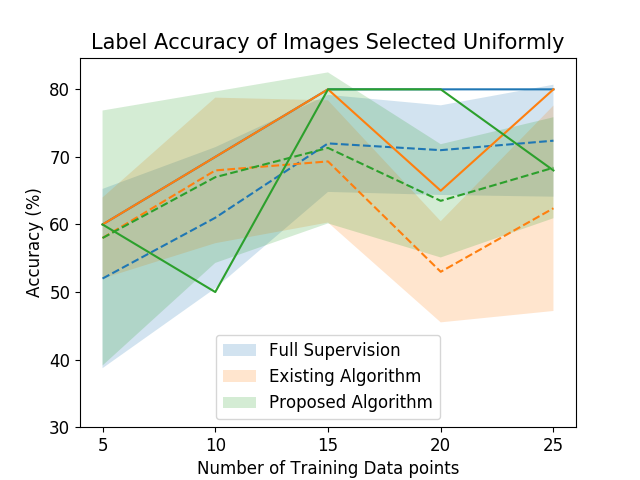}}
  \subfloat{
    \includegraphics[width=0.49\textwidth]{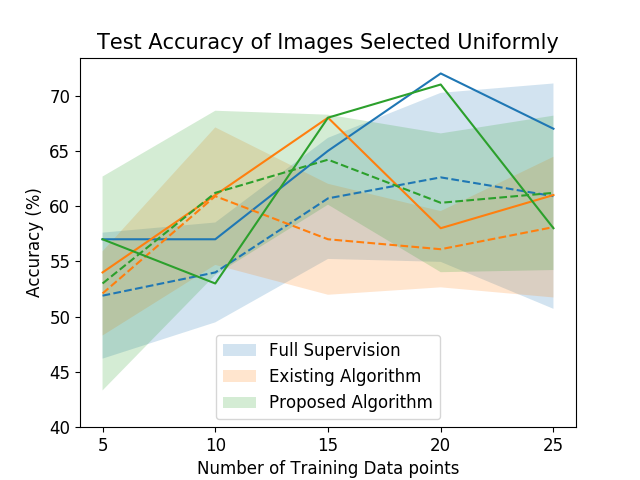}}
  \caption{Performance on uniformly selected data points.}
  \label{fig:uniform}
\end{figure}

Figure \ref{fig:kmed} and Figure \ref{fig:uniform} show the label accuracies and test accuracies for results of $25$ greedily selected medoids and $25$ uniformly selected data points, respectively.
The test accuracy measures the performance of each classifier learnt from inferred labels on a test dataset of size $100$, which is uniformly selected without replacement after the selection of training data points.
For label accuracies, we calculated the scores for each trial.
For test accuracies, we uses aggregated results and calculated only once.
The mean value from results of $10$ annotators are shown in dashed lines and the standard deviation are shown by the shadow. The value from aggregated results are shown in solid lines.
The proposed algorithm showed competitive performance to fully supervised learning \textit{without accessing explicit class labels at all}.

When increasing the number of training data, we observed the proposed algorithm could also show stable and promising generalization ability competitive with full supervision.
However, the performance of the existing algorithm \cite{xu17} was not stable, because it separated data points into small bags, and queried a random subset of each bag for \textit{explicit class labels}.
With fewer training data, the size of each bag was small and it could query most of a bag for explicit class labels, thus achieved high labeling accuracy.
However, with more training data, a reasonable budget restrained the size of the subset from each bag for querying explicit class labels, thus resulting the drop in performance.

Then we analysed the properties of pairs selected for $O_2$.
Different from last paragraph, these pairs were selected by the algorithm ran on crowdsourcing results.
We introduce another type of difficulty: \textit{pair difficulty} for a pair of data points.
We investigated the relationship between pair types and pair difficulties.
The user evaluation of \textit{pair difficulties} were $0.16\,(\pm0.33)$, $0.17\,(\pm0.10)$ and $0.56\,(\pm0.09)$, respectively.
It matches the intuition that annotators confused when both images were difficult to classify.

\begin{figure}[htbp]
  \centering
  \includegraphics[width=0.8\textwidth]{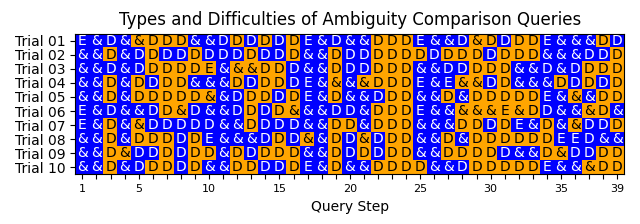}
  \caption{Uncertainty comparison query types and difficulties.}
  \label{fig:query}
\end{figure}

Figure \ref{fig:query} shows the trajectories of actually queried uncertainty comparisons of $10$ trials, indicating \textit{easy pairs} by white and \textit{difficult pairs} by orange.
Note that each query consisted of a pair of images.
Taking Trial $02$ as an example, we observe that for the first query, an annotator found it easy to assign the explicit class label to one image and difficult for the other. This also holds for the second query. The same annotator then found it difficult to assign explicit class labels to both images in the third query.
Another annotator found it easy to compare uncertainties than explicit labeling images in the first and second queries, and difficult for the third query.
We can observe that more \textit{difficult pairs} are queried on the latter half of the executions.
This can be interpreted that the algorithm successfully separated difficult data from easy data at an early stage.
Note that for the purpose of separating data by different difficulties,
the results of `E' pairs and `D' pairs do not effect too much as the data points in these pairs have similar difficulty.

\begin{figure}[htbp]
  \centering
  \includegraphics[width=0.8\textwidth]{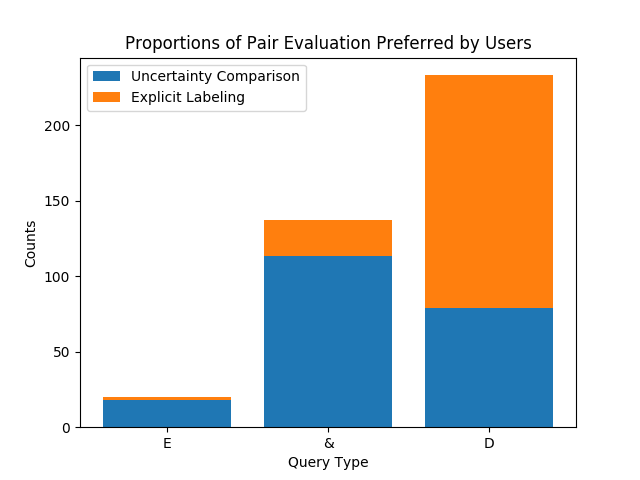}
  \caption{Histogram of uncertainty comparison query types and difficulties.}
  \label{fig:query-hist}
\end{figure}

Figure \ref{fig:query-hist} shows the corresponding histogram.
As pair type becoming difficult, the proportion of pairs evaluated as difficult for uncertainty comparison increased as expected.
Although blue areas are more preferable than orange areas, the proposed algorithm is not significantly influenced by the orange proportion of `D' pairs.

\paragraph{User comments}
At the end of each questionnaire, we also asked annotators to answer their opinions on these tasks in free text.
We select some of representative opinions and list their English translation \footnote{The translation is based on the results of DeepL (https://www.deepl.com/translator).}.

The following list shows advantages of positivity comparisons over explicit labeling.
\begin{itemize}
  \item It is easy to choose between ``NA" or ``WO" even if you can't read the word.
  \item You can choose the one you can easily recognize.
  \item You can choose the letters by your feeling.
  \item Unlike direct judgments, there is no clear correct answer, so it is possible to create questions that are easy for anyone to answer.
  \item When it's not too curled up, it's easy to choose.
\end{itemize}

The following list shows disadvantages of positivity comparisons over explicit labeling.
\begin{itemize}
  \item If you cannot read either of them, your selection criteria will be blurred.
  \item It is hard to judge a flaw when it's curled up.
  \item It is not sure if the decision is accurate.
  \item You need to stop and compare both images carefully, and may feel a great sense of hesitation before making a decision.
  \item Unlike direct judgement, there is no clear correct answer, and if neither letter is difficult to judge, you don't have to think about the answer. You can make a good choice.
\end{itemize}

The following list shows advantages of uncertainty comparisons over explicit labeling.
\begin{itemize}
  \item It's easy to choose if you can read one or the other somehow.
  \item It's quick and intuitive and I understand it quickly.
  \item Can be narrowed down if both are recognized as ``NA" or ``WO".
  \item It's easy to imagine how easy it is to read by just the simple criterion of being able to read, and how easy it is to read by pronouncing it in your head.
  \item It is highly flexible and does not have any restrictions.
\end{itemize}

The following list shows disadvantages of uncertainty comparisons over explicit labeling.
\begin{itemize}
  \item You can only seem to read them, but you can't tell whether you actually chose the correct answer or not.
  \item I don't know if other people can quickly recognize.
  \item If the words are not read as ``NA" or ``WO", I use the elimination method to select.
  \item When neither of them is likely to be readable, I tend to choose them at random.
  \item Unlike direct judgments, there is no clear correct answer, which makes it difficult to evaluate the competence of the annotator.
\end{itemize}

As we can see from above lists, it is difficult to choose when both images in a pair are not recognizable.
This may affect the accuracy of the existing method, as it is required to sort the whole dataset.
However, this does not significantly downgrade the performance of the proposed algorithm, as either one in the pair satisfying the desired uncertainty.
Moreover, it is interesting to see the various criterion used by annotators.

\subsection{Car preference task}
The pairwise positivity oracle $O_1$ is extensively used in preference learning.
Thus in this study, we used a car dataset \cite{car} to simulate a binary classification using \textit{user preference}, denoting car images a user likes as positive and those a user dislikes as negative.
Note the true labels differ for each user, as different users may have different preferences for cars.

\paragraph{Goals}
We want to verify if the proposed comparison oracle is useful for binary classification of individual user preference.

\paragraph{Method}
In this user study, we conducted crowdsourcing in two ways.
\begin{itemize}
  \item
    First, we collected user preference by five-stage evaluation.
    Stage one indices the user likes the car very much and stage five indices the opposite.
    This can be seen as different ranges of $p(y=1|x)$ for a given image $x$, thus can be used to simulate both pairwise comparison oracles.
    For eliciting explicit labels, we considered the first two stages as positive.
  \item
    Second, we directly collected user feedback of two kinds of pairwise comparisons on all possible pairs for a fixed set of training data.
\end{itemize}
We used an interface that is similar to the one used in the first user study.

\paragraph{Data}
The original dataset \cite{car} consists of $196$ categories.
We trained a ResNet18 \cite{resnet} model for classifying car categories to extract useful features.
Based on extracted features, we greedily selected a single medoid for each class to collect $196$ images.
For the first crowdsourcing task on five-stage evaluation, we then uniformly selected $150$ images.
For the second task, we greedily selected $25$ medoids based on extracted features for training and used the left $125$ images for testing.
We collected user feedback of all possible $300$ pairs for both kinds of pairwise comparisons.
All tasks are answered by four users.
After inferring labels, we trained both a neural network classifier and a $k$-NN classifier for each setting.

\paragraph{Algorithm feasibility using simulated pairwise comparisons}
Using five-stage evaluation to simulate pairwise comparisons, we had the freedom of choosing various sets of training data points.
Thus, we conducted experiments with different sizes of training data points that are selected either uniformly or greedily as medoids using extracted features.
The simulated feedback was noisy in the sense that when two images has same stage evaluation, we can only randomly answer one with equal probability.

\begin{figure}[htbp]
  \centering
  \includegraphics[width=0.95\textwidth]{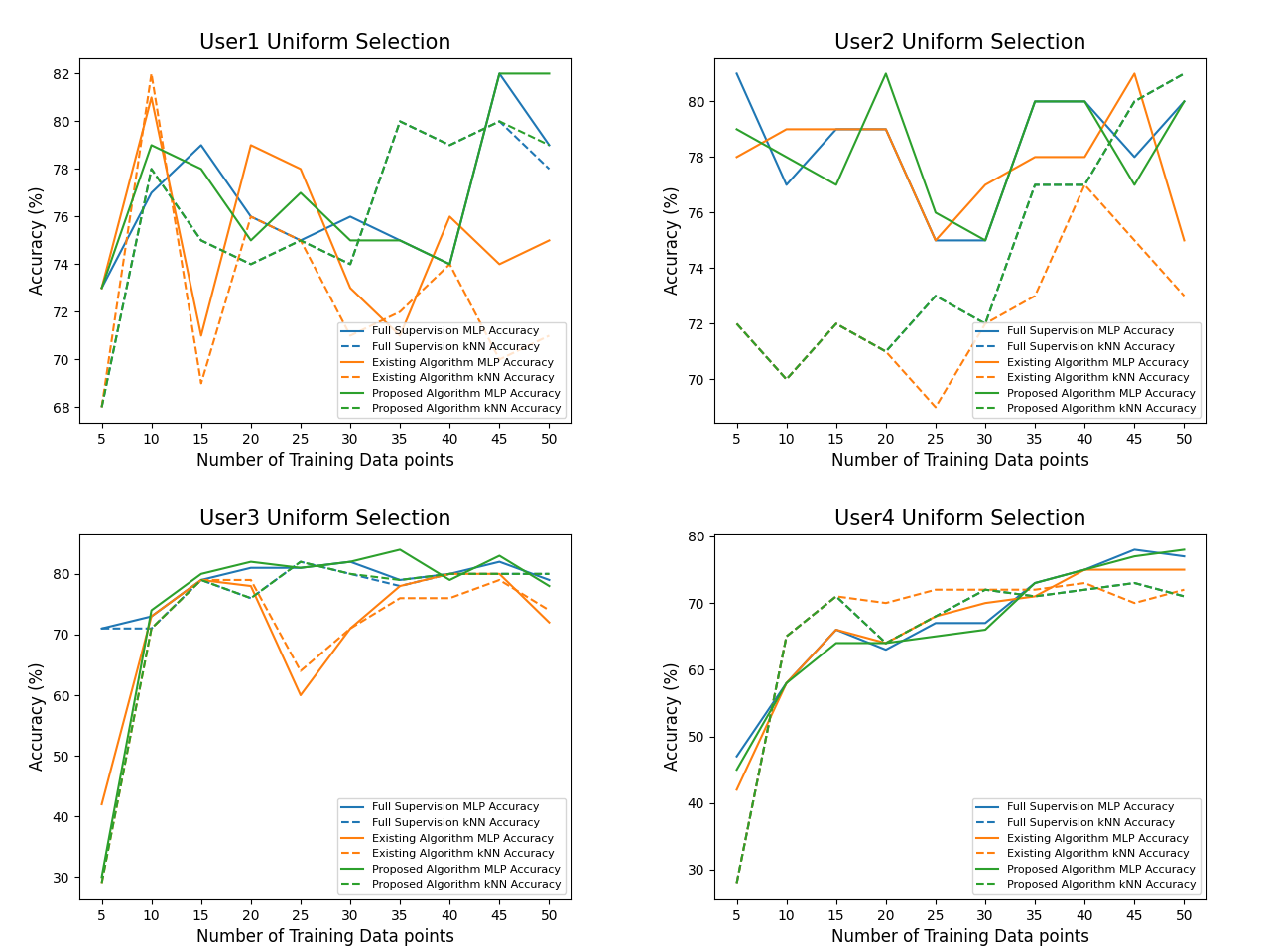}
  \includegraphics[width=0.95\textwidth]{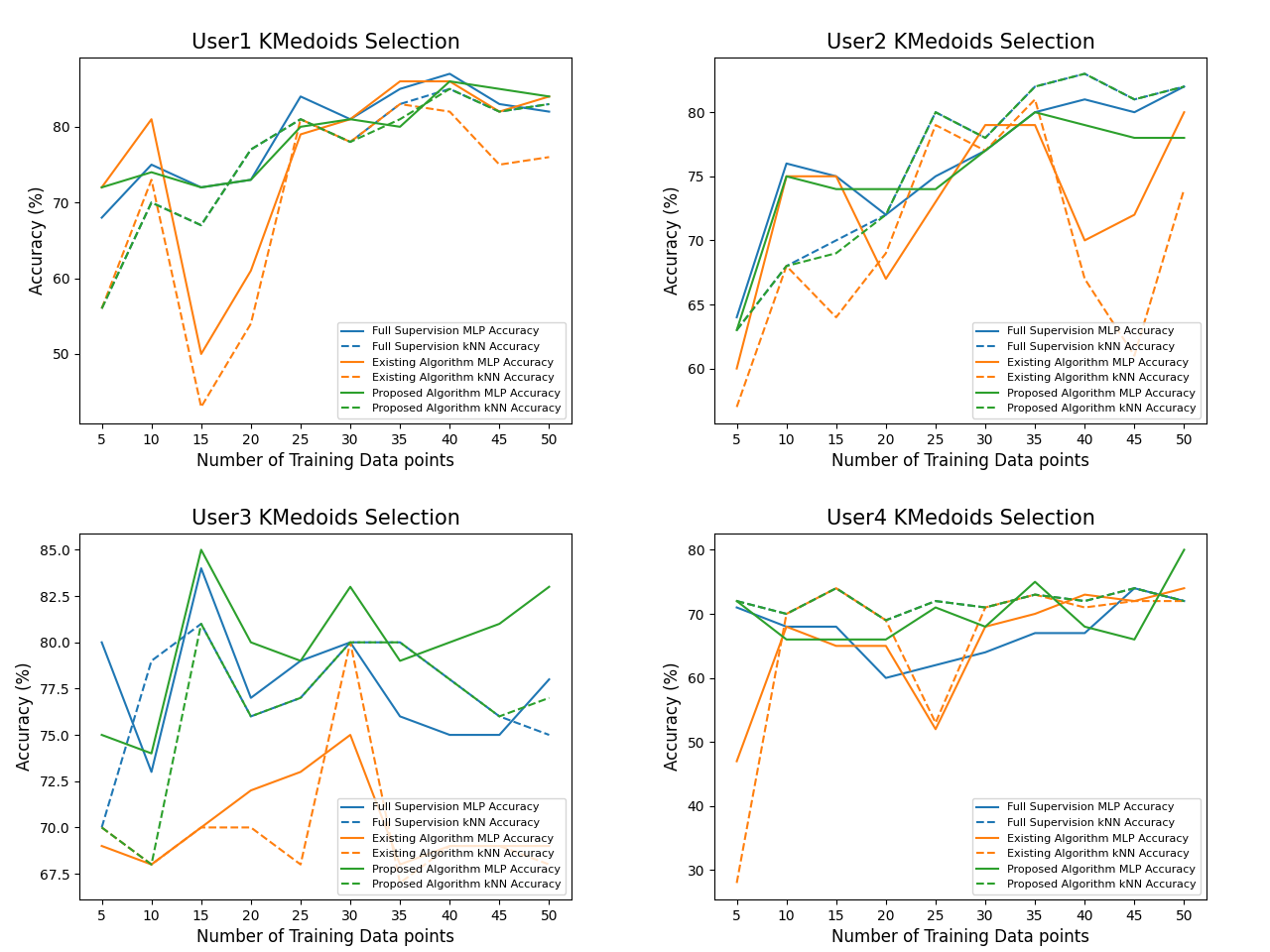}
  \caption{Test accuracies using simulated pairwise comparisons.}
  \label{fig:car-sim}
\end{figure}

As shown in Figure \ref{fig:car-sim}, the proposed method using only simulated pairwise comparisons showed competitive performance to fully supervision.
The performance of the existing method was not stable, because the quick sort subroutine is very sensitive to the results of pairwise comparison, which could be random in this case.
However, the proposed algorithm showed consistent performance under the same situation.

\paragraph{User Comments}
After a user finished answering all questions, we asked comments on the following open questions.
The answers below are summaries of comments from four users.

Question $1$: What are the characteristics of pairs that are easy for preference uncertainty comparison.
\begin{itemize}
  \item When one of the car falls in the middle of like and dislike, or falls in a preferred category.
  \item When two cars are completely different from each other.
\end{itemize}

Question $1$: What are the characteristics of pairs that are difficult for preference uncertainty comparison.
\begin{itemize}
  \item When two cars have similar appearance or preference.
\end{itemize}

Question $3$: What factors decide the difficulty of preference uncertainty comparison.
\begin{itemize}
  \item Appearance; category; experience.
\end{itemize}

Question $4$: What other items that preference uncertainty comparison may work?
\begin{itemize}
  \item Food; plants; shoes; cloth; things that are unusual in daily life.
\end{itemize}

Question $5$: What other measures other than preference uncertainty comparison may work?
\begin{itemize}
  \item Fairness; measures that everyone is familiar with; measures that based on experience.
\end{itemize}

\section{Conclusion}
In this paper, we address the problem of interactive labeling and propose a novel uncertainty comparison oracle, followed by a noise-tolerant theoretical-guaranteed labeling algorithm without accessing explicit class labels at all.
We then confirm the performance of the algorithm theoretically and empirically.
For future work, eliminating $\mathcal{O}(n)$ from one of the query complexity can improve the efficiency.
On the other hand, extending the uncertainty comparison oracle to multiple data points and multiple classes is a promising direction.

\section*{Broader Impact}
We believe this research will benefit researchers in all fields who are seeking for a more effective and less laborious annotation method for their unlabeled datasets.
It can foster applications of machine learning by lowering the annotation barrier for people without specific professional knowledge.
It can also benefit domain experts with professional knowledge by saving their time for more important tasks.
Furthermore, collecting comparison information can potentially mitigate annotation biases of explicit labeling.
It can also serve the aim of protecting privacy by not querying the explicit class labels in some cases.

For the negative side, it may harm the performance of downstream classification models when the comparison annotation is mostly incorrect.
However, there would be no consequential ethical issues of failure of the method.

\bibliographystyle{plain}
\bibliography{main}
\clearpage

\appendix
\section{Proof of Theorem \ref{thm:proposed}}
\label{supp:proposed}
\begin{proof}
The Algorithm \ref{alg:proposed} consists of two steps: selection of relatively uncertain points and assigning labels by majority vote.

For the first step, the algorithm of Mohajer et al. \cite{mohajer17} is executed using parameters $K=t$ and $m$.
By adapting Theorem 1 of Mohajer et al. \cite{mohajer17},
we know that if $m \geq \frac{C_1\max(\log\log{n}, \log{t})}{(0.5-\epsilon_2)^2}$,
then the correct top-$t$ points can be identified with probability at least $1-{\log{n}}^{-C_2}$.

For the second step, we analyze the probability that a point $x \in D \setminus D'$ is correctly inferred.
Without loss of generality, we assume the correct label for $x$ is $1$ and we calculate the probability that $\sum_{x_j \in D'} O_1(x, x_j) \geq \frac12$.

Let $Z_j \triangleq O_1(x, x_j)$ denotes the random variable representing the outcome of every call to oracle $O_1$.
Because $D'$ is assumed to be correctly identified,
so $p(y|x) \geq p(y|x_j)$ for every $x_j \in D'$,
thus the expectation of $Z_j$ is $1-\epsilon_1$.
Also note that $Z_j$ only takes a value of either $0$ or $1$,
thus by applying Hoeffding's inequality to $Z_1, Z_2, \cdots, Z_t$,
we have
\begin{equation}
 \mathrm{Pr}\left[ \frac1t\sum_{j=1}^t Z_j - (1-\epsilon_1) \leq -(0.5-\epsilon_1) \right] \leq \exp\left(-2t(0.5-\epsilon_1)^2\right).
\end{equation}
This actually expresses the probability that $\frac1t\sum_{j=1}^t Z_j$ is smaller than $0.5$.

Let $a\triangleq\exp\left(-2t(0.5-\epsilon_1)^2\right)$.
Because $t$ is selected so that $a\leq\frac12$
and $\frac1t \sum_{j=1}^t Z_j$ is bounded within $[0, 1]$,
therefore for a single $x \in D \setminus D'$ it holds that
\begin{align}
 \mathrm{Pr}\left[\frac1t\sum_{j=1}^t Z_j \geq \frac12 \right] & \geq 1 - a \\
 & \geq \exp(-a(a+1)).
\end{align}

For points in $D'$,
because we assign random labels,
there is positive probability that all assigned labels are wrong.

In conclusion,
for all points in $D \setminus D'$ correctly labeled,
the error rate $\epsilon = \frac{t}{n}$ can be achieved with probability at least $1-\delta$
where $\delta \triangleq 1 - (1-\log{n}^{-C_2})\exp(-a(a+1)(n-t))$.

For query complexities, as $O_1$ is queried $t(n-t)$ times,
the query complexity of $O_1$ is $\mathcal{O}\left(\frac{n}{\epsilon_1^2}\right)$.
Moreover, as indicated by Eq.~(17) of \cite{mohajer17},
the query complexity of $O_2$ is $\mathcal{O}\left(\frac{n\log\log{n}}{\epsilon_2^2}\right)$.
\end{proof}

\section{Proof of Theorem \ref{thm:knn}}
\label{supp:knn}
\begin{proof}
First, we bound the difference between $\hat{f}(x;k)$ and $f(x)$.
Similar to Reeve et al. \cite{reeve19}, we define
$\tilde{f}(x;k)=\mathbb{E}_{p(y|x)}=\frac1k\sum_{q=1}^k y_{\tau_q(x)}$.

Then we have
\begin{equation}
 \left|\hat{f}(x;k) - f(x)\right| \leq \left|\hat{f}(x;k) - \tilde{f}(x;k)\right| + \left|\tilde{f}(x;k) - f(x)\right|.
\end{equation}
For the first term in RHS, from Theorem \ref{thm:proposed},
we know it is bounded by $\frac{2\epsilon}k$ with probability at least $1-\delta$.
For the second term in right hand side,
from Lemma 4.1 in \cite{reeve19},
we have it is bounded by $\omega\left(\frac{2k}{n}\right)^\lambda$
with probability at least $1-\delta'$ for $\delta'>0$ and $\frac{n}2\geq k\geq 4\log(\frac1{\delta'})+1$.
Thus combing the two inequalities,
we have the left hand side is bounded by
$\Delta \triangleq \frac{2\epsilon}k + \omega\left(\frac{2k}{n}\right)^\lambda$
with probability at least $(1-\delta)(1-\delta')$.
This means with at least the same probability,
a randomly drawn point from $\mathcal{X}$ will fall in the set
$$\mathcal{X}'\triangleq\{x\in\mathcal{X}:|\hat{\eta}(x)-\eta(x)|\leq\Delta\}.$$

Thus
\begin{align*}
& R(\hat{f}) - R(f^*) \\
= & \int_\mathcal{X} \left|\eta(x)-\frac12\right| \mathbbm{1}_{\hat{f}(x) \neq f^*(x)} d\mu(x) \\
= & \int_{\mathcal{X}'}\left|\eta(x)-\frac12\right| \mathbbm{1}_{\hat{f}(x) \neq f^*(x)} d\mu(x) \left(\text{with probability at least} (1-\delta)(1-\delta') \right)\\
\leq & \int_{\mathcal{X}} \left|\eta(x)-\frac12\right| \mathbbm{1}_{\left|\eta(x)-\frac12\right|\leq\Delta} d\mu(x) \\
\leq & C\Delta^{\alpha+1}.
\end{align*}
\end{proof}

\section{Proof of Corollary \ref{thm:a2}}
\label{supp:a2}
\begin{proof}
Similar to the approach in Xu et al. \cite{xu17},
we use induction to show that at the end of every step $i$,
$\mathbb{E}_{\p_\X}[h(x)\neq h^*(x)]\leq4\epsilon_i$ always holds with probability at least $(1-\delta)^{\log(\frac1\epsilon)}$ for a universal $\delta$, which is obvious for $i=0$.

Then, with a little abusing of notations, we have
\begin{align*}
|{x\in S_i:h(x)\neq h^*(x)}| & = |{x\in D_i:h(x)\neq h^*(x)}| \\
& \leq |{x\in D_i:h(x)\neq \hat{y}}| + |{x\in D_i:h^*(x)\neq \hat{y}}| \\
& = 2\epsilon_i|S_i|.
\end{align*}

Thus $\text{P}_{x\sim S_i}[h(x)\neq h^*(x)] = \frac{|{x\in S_i:h(x)\neq h^*(x)}|}{|S_i|} \leq 2\epsilon_i$.
Having $c_0\in(1,\infty)$ and $\gamma\in(0, 1)$,
using Lemma 3.1 from Hanneke et al. \cite{hanneke14},
we have $\text{P}_{x\sim\p_\X}[h(x)\neq h^*(x)]\leq4\epsilon_i$
with probability at least $1-\gamma$,
providing $c_0\frac{d\log(\frac{|S_i|}d)+\log(\frac1\gamma)}{|S_i|}\leq\epsilon_i$.
Setting $\gamma=1-(1-\delta)^{\log(2\epsilon)}$,
We have $\text{P}_{\p_\X}[\hat{h}(x)\neq h^*(x)]\leq\epsilon$ with probability at least $(1-\delta)^{\log(\frac1\epsilon)}(1-\delta)^{\log(2\epsilon)} = 1-\delta$ at the end of the algorithm.
\end{proof}

\end{document}